\documentclass[11pt,a4paper]{article}
\usepackage[hyperref]{acl2021}
\usepackage{times}
\usepackage{latexsym}
\usepackage{soul}

\usepackage{microtype}

\aclfinalcopy % Uncomment this line for the final submission
\usepackage{amsfonts}
\usepackage{url}
\usepackage{xcolor}
\usepackage[ruled,linesnumbered]{algorithm2e}
\usepackage{graphicx}
\usepackage{caption}
\usepackage{subcaption}
\usepackage{bbm}

\usepackage{amsmath,bm}

\usepackage{microtype}

\title{VLM: Task-agnostic Video-Language Model Pre-training\\for Video Understanding}

\date{}

\author{Hu Xu$^{1}$, Gargi Ghosh$^{1}$, Po-Yao Huang$^{12}$, Prahal Arora$^{1}$, Masoumeh Aminzadeh$^{1}$ \\ 
\textbf{Christoph Feichtenhofer}$^{1}$, \textbf{Florian Metze}$^{1}$ \and \textbf{Luke Zettlemoyer}$^{1}$ \\
$^1$Facebook AI\\
$^2$Carnegie Mellon University\\
\texttt{\{huxu,gghosh,berniehuang,prarora,masoumeha,}\\
\texttt{feichtenhofer,fmetze,lsz\}@fb.com}
}

\begin{document}
\maketitle
\begin{abstract}
%Multi-modal pre-training attempts to learn robust task-agnostic embeddings, which contain the information contained in short video clips, and which can be fine-tuned towards downstream tasks such as video retrieval. 
%Existing methods either adopt a single encoder that requires both modalities, limiting their use for retrieval-style downstream tasks, or more complex schemes with two unimodal encoders, limiting early cross-modal fusion.
%We present a new, simplified multimodal pre-training approach based on a single masked language model that can accept either video or text input, or both.
%We introduce a novel masking scheme that improves mixing across modalities (e.g.~by forcing masks for text to predict the closest video embeddings), while also maintaining separability (e.g.~unimodal predictions are sometimes required, without using all the input). 
%Experimental results show strong performance across a wider range of tasks than previous works, often outperforming task-specific techniques.

We present a simplified, task-agnostic multi-modal pre-training approach that can accept either video or text input, or both for a variety of end tasks.
Existing pre-training are task-specific by adopting either a single cross-modal encoder that requires both modalities, limiting their use for retrieval-style end tasks or more complex multitask learning with two unimodal encoders, limiting early cross-modal fusion.
We instead introduce new pretraining masking schemes that better mix across modalities (e.g. by forcing masks for text to predict the closest video embeddings) while also maintaining separability (e.g. unimodal predictions are sometimes required, without using all the input).
Experimental results show strong performance across a wider range of tasks than any previous methods, often outperforming task-specific pre-training. Code is made available at \url{https://github.com/pytorch/fairseq/tree/main/examples/MMPT}. 
%In general, we show it is possible to fine-tune our model with different types of attention masks to meet the needs of diverse downstream tasks.
%\todo{make it one}.
%This further allows for learning a shared vision-language space end-to-end on transformers.
%Experimental results show that this baseline has competitive or better performance while requiring a significantly smaller number of parameters.

\end{abstract}

\section{Introduction}
\label{sec:intro}
%Self-supervised learning (SSL) achieves remarkable performance gains by harvesting a huge amount of training examples from unlabeled data via proxy tasks.

% aims to perform pre-training once and uses a general model and proxy tasks to cover a wide range of downstream tasks with less assumption on which downstream task the model will be fine-tuned later.
% As a result, we observe that multimodal pre-training have the following challenges:
% (1) unlike LMs, the design choices of architecture are unclear given multiple modalities from different streams of data and their interactions.
% (2) the proxy task faces more challenges to cover popular downstream tasks such as text-video retrieval that are unusual for pre-trained LMs\cite{devlin-etal-2019-bert}.

% In multimodal learning, however, task-agnostic pre-training is seemingly harder, probably due to the complexity of modalities and diverse needs of downstream tasks.
% For example, text-video retrieval (in joint latent space) is a popular downstream task for Howto100M dataset~\cite{miech2019howto100m}. Existing studies either limit such problem to a (pair-wise) matching problem (in metric space) \cite{sun2019videobert,zhu2020actbert}, or go for task-specific pre-training 
% \cite{miech2019howto100m,miech2020end,li-etal-2020-hero,luo2020univilm}.
%(e.g., retrieval only\cite{miech2019howto100m,miech2020end} or multitask learning \cite{li-etal-2020-hero,luo2020univilm}).

We study the challenge of achieving \textit{task-agnostic} pre-training for multimodal video understanding, building on recent unimodal approaches such as pretrained language models for text~\cite{peters-etal-2018-deep,devlin-etal-2019-bert}.
%and contrastive learning for images~\cite{chen2020simple,he2020momentum}.
Although certain language models are near task-agnostic~\cite{devlin-etal-2019-bert,lewis-etal-2020-bart} on NLP tasks,
being task-agnostic on multi-modal tasks are more challenging due to cross-modal tasks such as text-video retrieval.
Existing video-and-language pre-trainings are task-specific, which adopt either (1) a cross-modal single encoder  \cite{sun2019videobert,sun2019contrastive,zhu2020actbert} favoring tasks that require cross-modal reasoning (e.g. video captioning), 
%that requires bi-modal input and mixes modalities for generation-style tasks;
or (2) multiple unimodal encoders/decoders \cite{miech2019howto100m,miech2020end,li-etal-2020-hero,luo2020univilm,korbar2020video} combining specific tasks that require separately embedding each modality (e.g. video retrieval). % for retrieval-style end tasks.
We instead show that it is possible to pre-train a task-agnostic model called video-language model (VLM) that can accept text, video, or both as input.
% MIL-NCE \cite{miech2019howto100m,miech2020end} uses retrieval-based pre-training and downstream tasks are also closer to retrieval-based tasks; 

\begin{figure}[t]
\centering    
\includegraphics[width=2.9in]{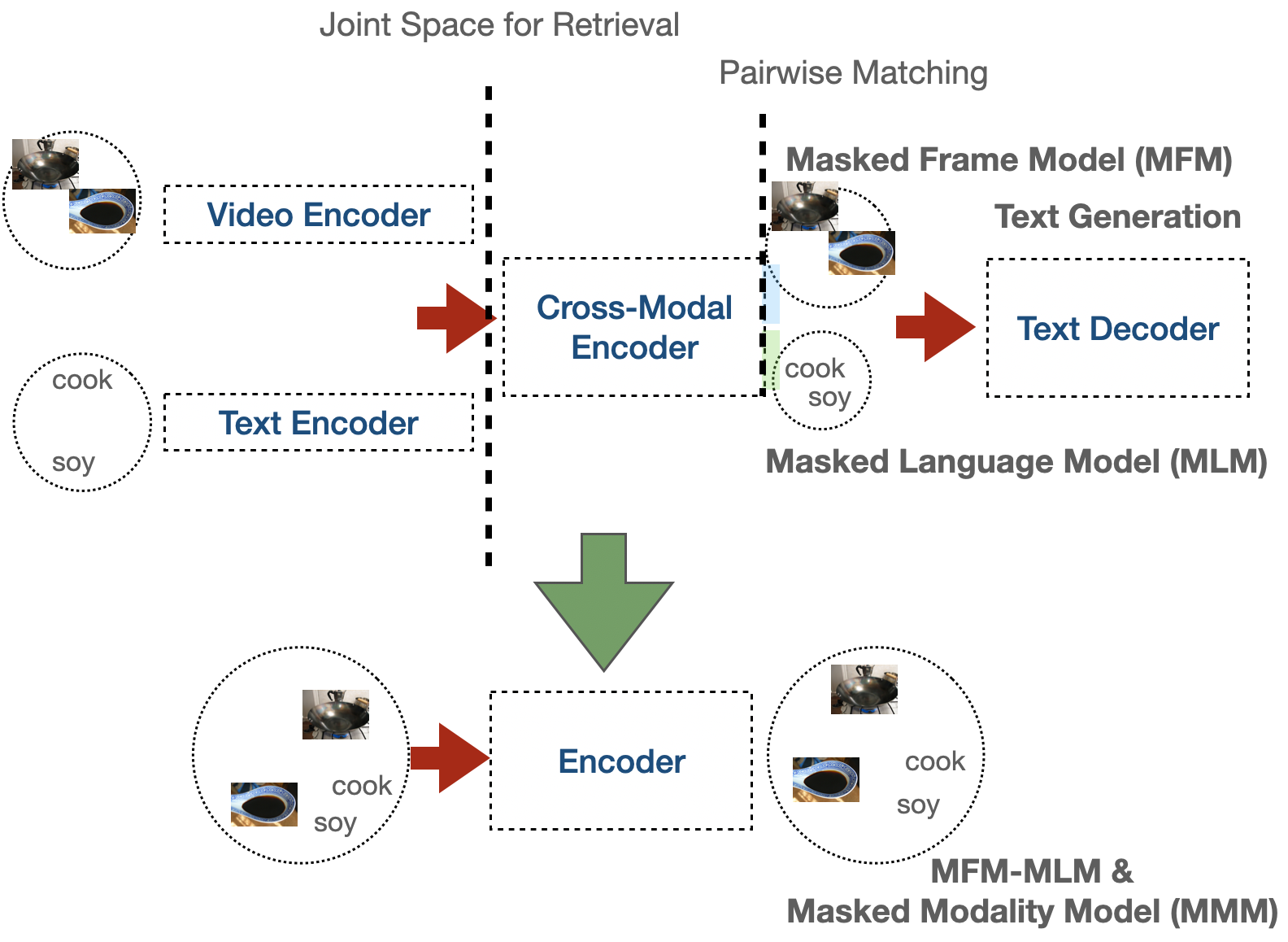}
    \caption{Existing models (upper figure) adopt complex architectures and multiple task-specific training to merge two streams of data to cover a wide range of downstream tasks (such as retrieval or text generation). Our video-language model (VLM) (lower figure) uses a single BERT encoder for task-agnostic pre-training (e.g. only masking tokens, no matching or alignment for specific end tasks) in a joint feature space, while still covering a wide range of tasks (see Figure~\ref{fig:ft}).}
\vspace{-4mm}
\label{fig:overview}
\end{figure}

As shown in Figure \ref{fig:overview},
this task-agnostic single encoder approach has several advantages:
(1) it reduces the complexity of pre-training with multiple losses and models (e.g. \citet{luo2020univilm}), and
(2) it holds less assumption on being close to end tasks as in retrieval-based pre-training \citet{miech2020end} and is as general as classic LMs, and
%where video/cross-modal encoders are randomly initialized and text encoders are from LMs.
% for example, the video encoder or cross-modal encoder are trained from scratch whereas the text encoder is tuned from an LM) and adopts multiple training stages to merge different encoders/decoders \cite{luo2020univilm};
(3) it encourages feature sharing among modalities when present, without sacrificing separability, and (4) it is more parameter efficient (see Section~\ref{sec:exp}, we achieve strong performance with $\text{BERT}_{\text{BASE}}$ sized models).
Table~\ref{tbl:num_param} summarizes the design choices of recent models. 

Our encoder is a transformer block that combines the existing masked frame model and masked language model (MFM-MLM) ~\cite{sun2019contrastive,li-etal-2020-hero,luo2020univilm}
% contrastive learning losses 
with two new methods to improve the learning of multi-modal fusion. %masking schemes. 
%We introduce two (2) fixes for pre-training:
% To accommodate a wide range of downstream tasks,
%(e.g., it is seemingly hard for \cite{sun2019videobert} to support text-video retrieval as in \cite{miech2019howto100m} so \cite{luo2020univilm,li-etal-2020-hero} adopts multi-task learning), 
First, we introduce a masking scheme called masked modality model (MMM) that randomly masks a whole modality for a portion of training examples (the rest of the examples goes for traditional MFM-MLM), thereby forcing the encoder to use the tokens from the other modality to produce tokens for the masked modality. 
%This encourages better cross model mixing 
%the issue of masked frame/language models (MFM/MLM) that may easily use tokens from a similar modality to produce the masked tokens.
We then introduce a single masked token loss to replace two (2) losses on video and text separately for MFM-MLM.
Masked token loss uses the embeddings of both video and text tokens to learn joint hidden states for the encoder.
% masking masked frame model (MFM) and masked language model (MLM) 
% into a single loss function.
%This fixes the losses of MFM and MLM on two separated modalities of negative examples of a limited number of tokens.

We also show it is possible to fine-tune a single encoder for a wide range of tasks by using task-specific attention masks. 
%This simplifies pre-training to be task-agnostic. 
Experiments demonstrate that it performs well on a wider range of tasks than previous models, including outperforming task-specific pre-training baselines with unimodal encoders of similar hyper-parameters by more than 2\% on retrieval tasks and 1\% on video captioning. 
Note that these results are also achieved with a much smaller model than previous approaches, further demonstrating the improved fusion and sharing across modalities. 
%We show that a single pre-trained encoder can out-perform solutions of unimodal encoders by 2-3 \% on retrieval tasks, and 1\% on video captioning.
% we reduce the complexity of downstream tasks to task-specific attention masks during fine-tuning (see Figure \ref{fig:ft}), while keep pre-training being task-agnostic.
% We further allow video and text tokens to share the same 
%encourage a fused 
% latent space 
% so vision and text are end-to-end fused on transformer \cite{vaswani2017attention}.
% We adopt masked frame model (MFM) and masked language model (MLM) in a fused token embedding space as proxy tasks and 
%and \todo{WIP, may drop MLM with a single loss in the end} explore to merge them as a single loss called masked token model.

In summary, the main contributions of this paper are as follows: (1) we propose to pre-train a task-agnostic encoder for video understanding;
(2) we introduce masked modality model (MMM) and masked token loss for cross-modal fusion during pre-training without sacrificing separability;
(3) experimental results show that the proposed simple baseline achieves competitive performance with significantly fewer parameters.

\section{Related Work}
\label{sec:related_work}
Numerous multimodal task-specific pre-training models are proposed for downstream visual-linguistic tasks.
In video and text pre-training, existing research adopts different design choices regarding proxy tasks and neural architectures for end tasks \cite{luo2020univilm}.

On one hand, 
%We summarize three pre-training paradigms to cover the previous vision-text pre-training model considering different encoder architecture in literature, as presented in Figure 2.
VideoBERT \cite{sun2019videobert}, Unicoder-VL \cite{li2020unicoder}, VL-BERT \cite{Su2020VL-BERT:}, UNITER \cite{chen2020uniter}, VLP \cite{zhou2018end},
ActBERT \cite{zhu2020actbert} adopt a \textit{shared} encoder approach, where the vision and text sequences are concatenated and input to a single Transformer\cite{vaswani2017attention} encoder. Although this approach is simple, it limits the types of downstream tasks to those that input both modalities simultaneously.
For example, \cite{sun2019videobert} may not be able to perform joint retrieval tasks and added another decoder for video captioning during fine-tuning.
\cite{zhu2020actbert} uses $\texttt{[CLS]}$ token for pairwise metric-learning based retrieval (which is an easier problem but requires a quadratic number of examples and is 50 times slower as reported in \cite{luo2020univilm}).

Meanwhile, many existing approaches adopt or add task-specific pre-training to accommodate retrieval and video captioning tasks (e.g. \textit{two-stream} encoders (video and text separately) and text decoders).
For example, \cite{miech2019howto100m,miech2020end,rouditchenko2020avlnet,ging2020coot,gabeur2020multi,alayrac2020self,patrick2021supportset,multiht100m_bernie} adopts a retrieval task for pre-training.
CBT \cite{sun2019contrastive}, HERO \cite{li-etal-2020-hero}, VideoAsMT \cite{korbar2020video} and UniVL \cite{luo2020univilm} adopt multi-task learning (MTL) to learn retrieval tasks on video and text encoders.
HERO \cite{li-etal-2020-hero} and UniVL \cite{luo2020univilm} adopts another cross-encoder to further learn the fusion of different modality.
UniVL \cite{luo2020univilm} and VideoAsMT \cite{korbar2020video} add another text decoder for video captioning.
Compared with the single-stream input in the shared encoder approach, two-stream encoders typically come with a complex architecture and proxy tasks to cover more end tasks.
To the best of our knowledge, none of the existing works target task-agnostic pre-training. 

\subsection{Image-Text Pre-training}
ViLBERT \cite{lu2019vilbert}, LXMERT \cite{tan-bansal-2019-lxmert} adopt two transformers for image and text encoding separately. 
VisualBERT \cite{li2019visualbert}, Unicoder-VL \cite{li2020unicoder}, VL-BERT \cite{Su2020VL-BERT:}, UNITER \cite{chen2020uniter}, Unified VLP \cite{zhou2020unified} use one shared BERT model.
These models employ MLM and pairwise image-text matching as pretraining tasks which are effective for downstream multimodal tasks.
Our fine-tuning for video captioning is inspired by Unified VLP \cite{zhou2020unified} that adopts attention masks and language model heads of BERT for image-captioning.

\subsection{Video-Text Pre-training}
VideoBERT \cite{sun2019videobert} and CBT \cite{sun2019contrastive} are the first works to explore the capability of pre-training for video-text. 
Although VideoBERT and CBT pre-train the model on multimodal data, the downstream tasks mainly take video representation for further prediction. 
ActBERT~\cite{zhu2020actbert} is a weakly-supervised pre-training method.
It leverages global action information to catalyze mutual interactions between linguistic texts and local regional objects and introduces a transformer block to encode global actions, local regional objects, and linguistic descriptions.
HERO \cite{li-etal-2020-hero} encodes multimodal inputs in a hierarchical fashion. Besides, two new pre-training tasks, video-subtitle matching and frame order modeling, are designed to improve representation learning. 
VideoAsMT \cite{korbar2020video} and UniVL \cite{luo2020univilm} further adopt a BART-style\cite{lewis-etal-2020-bart} text generation task for downstream tasks such as video captioning and UniVL adopts a EnhancedV training stage to mask all text tokens for better learning of generation. 

\section{Pre-training}
\label{sec:pretrain}

As a reminder, our goal is to train a  \textit{task-agnostic} model for various tasks in video-text understanding. 
%the goal of self-supervised learning is to turn unlabeled data into a huge amount of training examples via proxy tasks.
This section introduces task-agnostic proxies for pre-training. % used in this paper.
We first describe two masking schemes as a baseline: 
masked frame model (MFM) for video frames and masked language model (MLM) for text tokens \cite{sun2019contrastive,li-etal-2020-hero,luo2020univilm}. 
Then we
%Then we introduce MTM that unifies MFM and MLM as a single loss.
introduce masked modality model (MMM) that encourage to  %can be naturally combined with MFM and MLM to 
learn the representations of one modality from the other.
Lastly, we introduce masked token loss that 
% combines these 3 proxy tasks 
unifies losses on masked video and text tokens
as a single loss function.

\subsection{Vector Quantization and BERT}
Assume we have a clip $(v, t)$ sampled from a video, where $v$ and $t$ corresponds to video modality and text modality, respectively.
Since videos are signals in continuous space, we first extract token embeddings from raw videos.
We decode $v$ into frames and then feed them into a (frozen) video encoder $\text{Encoder}_\text{video}(\cdot)$ and a trainable MLP layer to obtain \textit{video tokens}:
% Then the hidden states of video frames from this video encoder are fed into an MLP layer.
% The output of the MLP layer will be in the same size as a text token (a.k.a word embeddings):
\begin{equation}
\begin{split}
\bm{x}_v = \text{MLP}(\text{Encoder}_\text{video}(\bm{f}_v)),
\end{split}
\end{equation}
where we use a bolded symbol to indicate a sequence and $\bm{f}_v$ is a sequence of continuous frames from a video. 
We use S3D~\cite{xie2018rethinking,miech2020end}, which is pre-trained via self-supervised learning on the Howto100M dataset.
The MLP layer allows the hidden size of video tokens to be the same as BERT's hidden sizes $d$: $x_v \in \mathbb{R}^d$.
Similarly, vectors for text tokens $\bm{x}_t$ are obtained via embedding lookup as in BERT.

To simplify multi-modal pre-training, we adopt a single BERT transformer with minimum changes.
% to learn single and cross-modality interactions 
% without complex vision/text and cross-modal
% encoders\cite{li-etal-2020-hero,luo2020univilm}. 
We first concatenate video tokens $\bm{x}_v$ and text tokens $\bm{x}_t$ via the \texttt{[SEP]} token so video and text belongs to one corresponding segment of BERT:
\begin{equation}
\begin{split}
\bm{x} = \texttt{[CLS]} \circ \bm{x}_v  \circ \texttt{[SEP]} \circ \bm{x}_t \circ \texttt{[SEP]}.
\end{split}
\end{equation}
We further mask $\bm{x}$ as $\bm{x}_\text{masked}$ (detailed in the next subsection) and feed the whole sequence into BERT:
%with a probability $m$ (by randomly set video tokens as zeros or text tokens as \texttt{[MASK]}) 
\begin{equation}
\begin{split}
\bm{h} = \text{BERT}(\bm{x}_\text{masked}),
\end{split}
\end{equation}
where $\bm{h}$ indicates the hidden states of the last layer of BERT.
To encourage learning video/text hidden states in a shared space for the masked token loss (introduced in Section \ref{sec:mmmmtl}), 
we use a \textit{shared} head to predict video/text token embeddings via a linear projection layer:
\begin{equation}
\begin{split}
e = \bm{W}h + b,
\end{split}
\end{equation}
where $e \in \mathbb{R}^d$ and $\bm{W}$ and $b$ are the weights from the prediction heads of BERT.
In this way, our model learns a joint embedding space for both video and text tokens from inputs to outputs of BERT.
This allows for pre-training a single encoder directly from any existing LMs and the only layer that requires initialization is the MLP layer.

\subsection{MFM-MLM}
Inspired by \cite{sun2019contrastive,li-etal-2020-hero,luo2020univilm}, we adopt masked frame model (MFM) for videos and masked language model (MLM) for text as a baseline.
Note that unlike LMs that typically come with a fixed vocabulary with a special $\texttt{[MASK]}$ token, video tokens are innumerable in the continuous space and we mask a video token by setting a video token with all zeros and ask the encoder to recover the video token.
via noisy contrastive estimation (NCE): 
%Instead, we construct NCE losses for MFM by encouraging 
% predict the masked video tokens being closer to original video token before masking and away from other video tokens within the same batch:
\begin{equation}
\begin{split}
\mathcal{L}_{\text{MFM}}= - \mathbb{E}_{s \sim V }\log\text{NCE}(x_s|\bm{x}_{\text{masked}}; V'),
\end{split}
\end{equation}
where $V$ is all indexes of video tokens and
\begin{equation}
\begin{split}
\text{NCE}(x_v|\bm{x}_{\text{masked}}; V') = \\
\frac{\exp(x_v^T e_v)}{\exp(x_v^T e_v) + \sum_{j\in V'} \exp(x_j^T e_v)},
% \frac{\exp(x_v^T e_v)}{\sum_{j\in V} \exp(x_j^T e_v)},
\end{split}
\end{equation}
where $V'$ indicates all non-masked video tokens within the same batch.
% Similar to BERT, the loss for MLM is defined as follows:
% \begin{equation}
% \begin{split}
% \mathcal{L}_{\text{text}}= - \mathbb{E}_{t \sim T} \log p(x_t|\bm{x}_{\text{masked}}).
% \end{split}
% \end{equation}
% Note that $\bm{x}_{\text{masked}}$ include both video and text tokens so a predicted hidden states $e_v$ (or $e_t$) can be inferred from both video and text tokens. 
The final loss is the sum of both MFM and MLM:
\begin{equation}
\begin{split}
\mathcal{L}_{\text{MFM-MLM}}=\mathcal{L}_{\text{MFM}} + \mathcal{L}_{\text{MLM}},
\end{split}
\end{equation}
where $\mathcal{L}_{\text{MLM}}$ is the same as BERT and we omit its details for brevity.
We experiment this classic baseline in Section \ref{sec:exp}.

% Although the losses of MFM and MLM can be summed to learn visual and text tokens, these two losses may potentially isolate video and text tokens into different sub-spaces.

% Instead, we construct NCE losses for MFM by encouraging the predicted (masked) video tokens being closer to original video token before masking and away from other video tokens within the same batch:
% \begin{equation}
% \begin{split}
% \mathcal{L}_{\text{vis}}= - \mathbb{E}_{v \sim V }\log\text{NCE}(x_v|\bm{x}_{\text{masked}}),
% \end{split}
% \end{equation}
% where $V$ is all indexes of video tokens and $\bm{x}_{\text{masked}}$ is the masked visual and text hidden states. $\text{NCE}(\cdot | \cdot)$ is defined as:
% \begin{equation}
% \begin{split}
% \text{NCE}(x_v|\bm{x}_{\text{masked}}) = \\
% \frac{\exp(x_v^T e_v)}{\exp(x_v^T e_v) + \sum_{j\in V} \exp(x_j^T e_v)},
% \frac{\exp(x_v^T e_v)}{\sum_{j\in V} \exp(x_j^T e_v)},
% \end{split}
% \end{equation}
% where $V$ indicates all non-masked video token indices (including $v$) within the same batch.
% Similar to BERT, the loss for MLM is defined as follows:
% \begin{equation}
% \begin{split}
% \mathcal{L}_{\text{text}}= - \mathbb{E}_{t \sim T} \log p(x_t|\bm{x}_{\text{masked}}).
% \end{split}
% \end{equation}
% Note that $\bm{x}_{\text{masked}}$ include both video and text tokens so a predicted hidden states $e_v$ (or $e_t$) can be inferred from both video and text tokens. 

% The final loss is the sum of MFM and MLM:
% \begin{equation}
% \begin{split}
% \mathcal{L}=\mathcal{L}_{\text{vis}} + \mathcal{L}_{\text{text}}.
% \end{split}
% \end{equation}

\subsection{MMM and Masked Token Loss}
\label{sec:mmmmtl}
\begin{figure}[t]
\centering
\includegraphics[width=3.in]{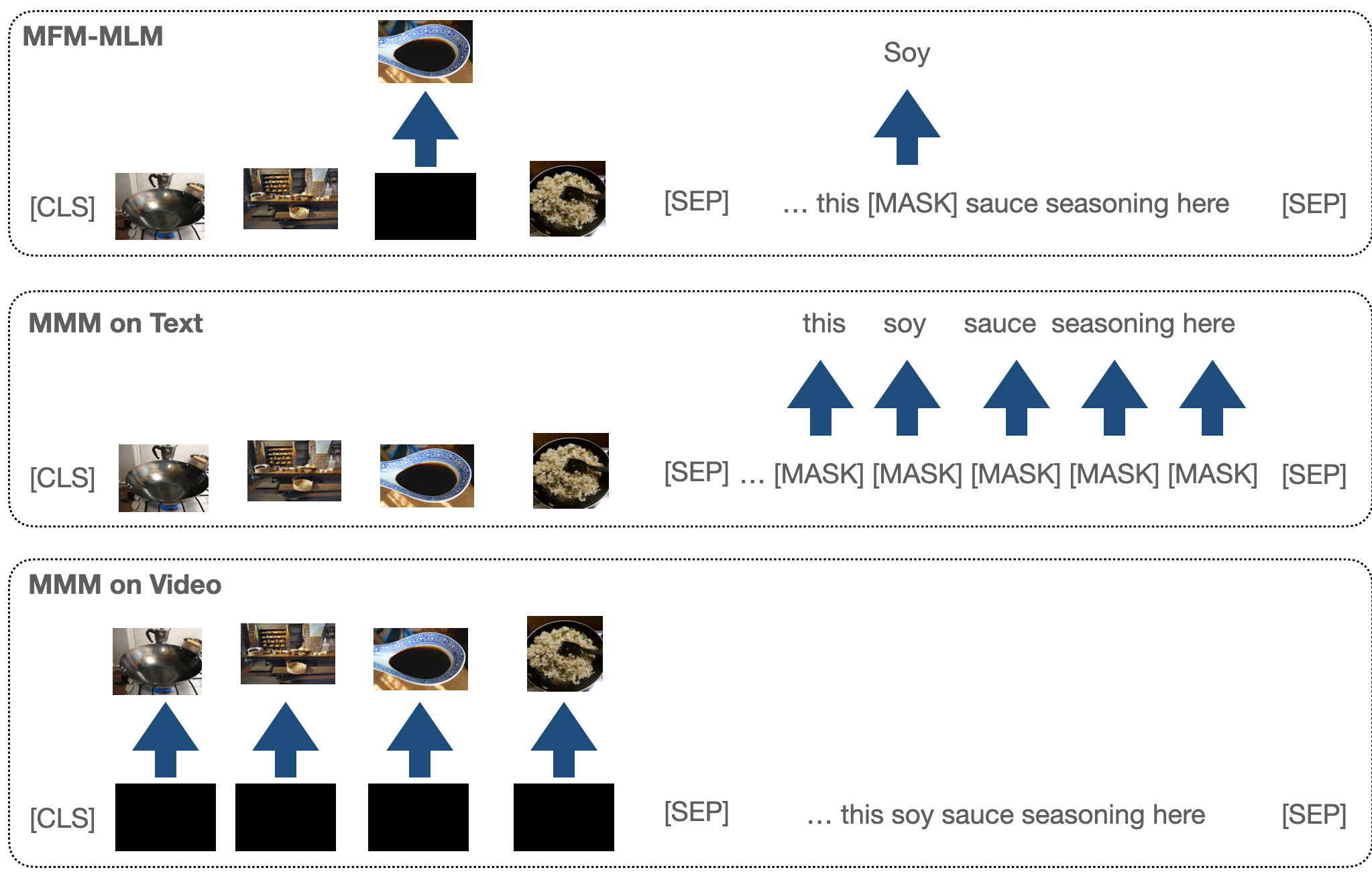}
    \caption{Task-agnostic pre-training (e.g. w/o task on retrieval-style alignment): MFM-MLM: 50\% of training examples are masked as masked frame model (MFM) and masked language model (MLM); the rest 50\% examples are masked as masked modality model (MMM) (25\% on text as in the second row and 25\% on video as in the third row).}
\vspace{-3mm}
\label{fig:task}
\end{figure}

\noindent \textbf{Masked Modality Model} We introduce masked modality modal (MMM) that masking either all video or all text tokens out for a given example of video-text clip. 
This masking scheme complements MFM-MLM (e.g. in our experiments 50\% of training examples are masked as MMM and the rest 50\% are masked as MFM-MLM).
This encourages the encoder to use tokens from one modality to recover the tokens for the other modality.
This resolves the issue that an encoder may use nearby tokens from their modality for prediction just because tokens from a single modality are closer
% (see Figure \ref{fig:token_embedding} in Appendix).
As in the lower two (2) sub-figures in Figure \ref{fig:task}, we either mask the whole modality of video or text so this modality can be ``generated'' from the other modality. 
Our experiments indicate that this is critical for pre-training a single encoder 
% be used later as unimodal encoders 
for retrieval tasks. %via isolated attention masks.

\noindent \textbf{Masked Token Loss} We further introduce masked token loss that unifies loss functions for MFM and MLM.
This loss encourages learning a joint token embedding space for video and text and both types of tokens contribute to the prediction of a masked (video or text) token.
This also improves the number of contrasted negative embeddings in two separate losses for MFM and MLM.

We define masked token loss $\mathcal{L}_{\text{VLM}}$ as the following:
\begin{equation}
\label{eq:lvlm}
\begin{split}
- \mathbb{E}_{s \sim V \cup D }\log\text{NCE}(x_s|\bm{x}_{\text{masked}}; V'\cup D_{\setminus s}),
\end{split}
\end{equation}
where $D$ is the word embeddings over the vocabulary of BERT and $D_{\setminus s}$ excludes token $s$ (if $s$ is a text token).
Further, $\text{NCE}(x_s|\bm{x}_{\text{masked}}; V' \cup D_{\setminus s})$ is defined as:
\begin{equation}
% \begin{flalign}
\begin{split}
\frac{\exp(x_s^T e_s)}{\exp(x_s^T e_s) + \sum_{j\in V'\cup D_{\setminus s}} \exp(x_j^T e_s)}.
% \frac{\exp(x_s^T e_s)}{\sum_{j\in V \cup T} \exp(x_j^T e_s)}.
\end{split}
\end{equation}
% \end{flalign}
Note that $j \in V' \cup D_{\setminus s}$ can be either a video or text token and one predicted token $e_s$ must be closer to the ground-truth token embedding (either a video token or word embedding) and be away from other embeddings of video/text tokens.
% This allows for more negative examples 
% than two losses on MFM and MLM 
We perform an ablation study in Section~\ref{sec:exp} to show that $\mathcal{L}_{\text{VLM}}$ works better than $\mathcal{L}_{\text{MFM-MLM}}$.  

\section{Fine-tuning}
\label{sec:ft}

\begin{figure*}[t]
\centering
\includegraphics[width=6.0in]{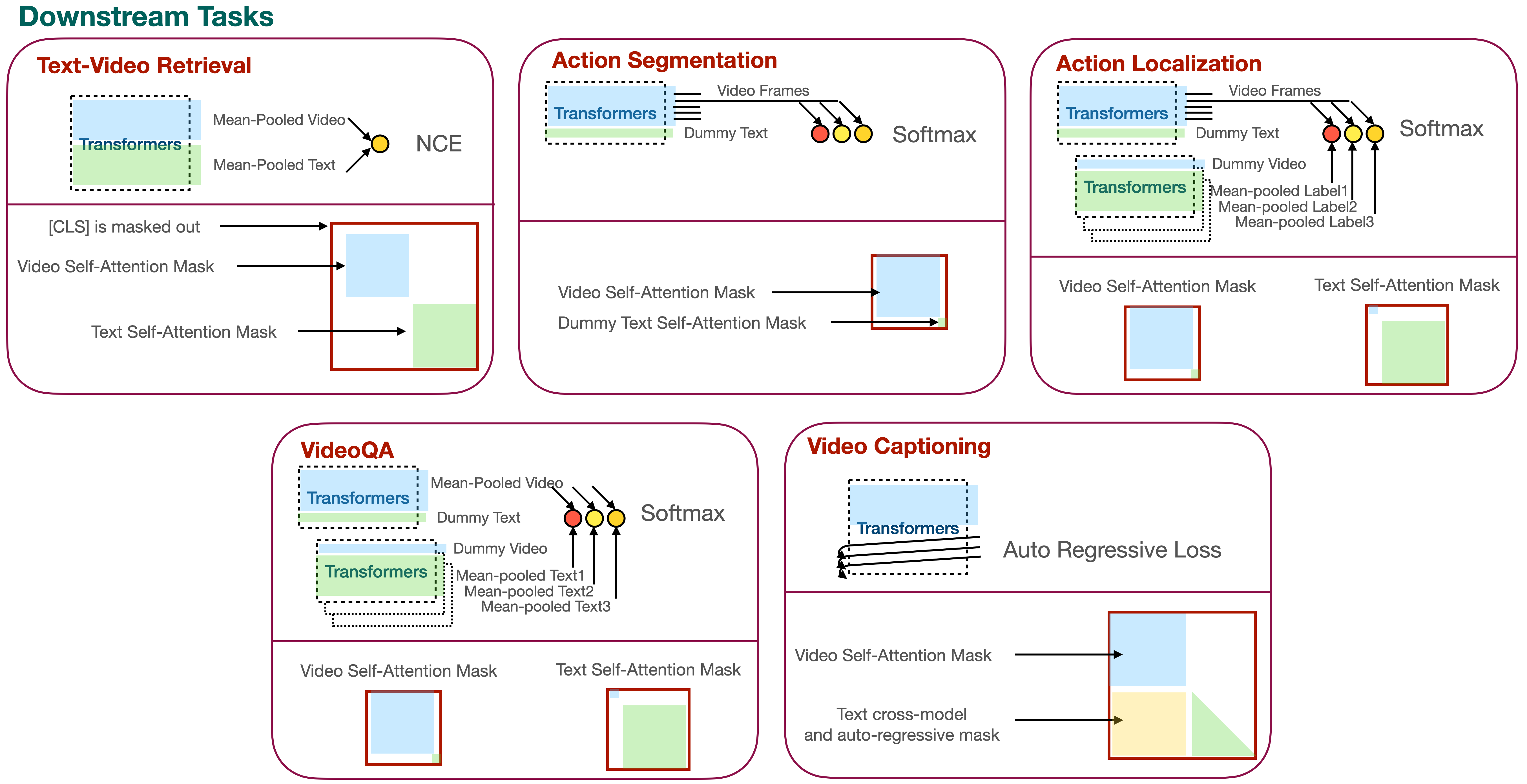}
    \caption{Fine-tuning of downstream tasks: we adopt different types of attention masks for BERT to accommodate downstream tasks that require different modalities: in each box, the upper sub-figure indicates a forward computation; the lower sub-figure indicates squared self-attention mask, where tokens from each row have a weighted sum of columns that are not in white colors.}
\vspace{-3mm}
\label{fig:ft}
\end{figure*}

In this section, we describe how to use different types of attention masks to fine-tune VLM for a variety of tasks, as shown in Figure \ref{fig:ft}.
%Our goal is not to exhaust all possible tasks VLM can support, but to demonstrate some hard tasks that are seemingly hard on a single encoder.

%\noindent \textbf{Text-Video Retrieval} 
\subsection{Text-Video Retrieval}
One major challenge of pre-training on a single encoder is how to adapt such a model to joint space retrieval without using unimodal encoders for task-specific pre-training on contrastive loss
(as in Howto100M~\cite{miech2019howto100m,miech2020end}). 
%Those tasks typically require two-stream encoders for vision and text respectively, where vision and text cannot see each other until reach a joint vision-text space.
%As a result, the hidden state of one text in such a joint space can have multiple similarities with the hidden states of multiple videos without forwarding quadratic numbers of text/video pairs as text/video matching in metric space as in \cite{zhu2020actbert}.
%The loss can be computed by contrasting a positive pair with other text/video negative pairs within a batch.
The main reason is that many existing models encode text and video tokens together via self-attention, and one cannot obtain hidden states for text/video alone.
%\todo{add this? } This is different from the metric-learning-based retrieval task that uses \texttt{[CLS]} token for quadratic computation between pairs of video and texts. 

To resolve this, we propose to apply an isolated attention mask with two squared masks that are diagonally placed, as shown in the lower sub-figure of the first box in Figure \ref{fig:ft}.\footnote{One can further reduce $O(m+n)^2$ complexity to $O(m^2+n^2)$ ($m$ and $n$ are lengths for video and text, respectively) by feeding video/text separately to BERT but we adopt squared masks for simplicity.}
These two squares disable video and text tokens to attend and see each other, while still allow video and text tokens to use the same self-attention layers for learning representations in the same feature space.
% This encourages learning features for computing similarity between video and text for retrieval tasks. %even though video and text tokens cannot see each other.
Further, note that the first and second \texttt{[SEP]} tokens of BERT will be used by video and text, respectively, aiming to learn sequence-level features\cite{clark2019does}. The \texttt{[CLS]} is disabled as no need to learn features across video and text.
After forwarding, all hidden states of video and text tokens are average pooled, respectively.
Then we use a contrastive loss on text-video similarity to discriminate a ground-truth video clip from other video clips in the same batch for a given text clip. 
% All similar/dissimilar pairs of similarity are computed from the hidden states of corresponding text and video. 
During the evaluation, to ensure video and text are isolated (to avoid leaking ground-truth of a similar pair), we split text and video and forward them separately.
We report an ablation study in Section~\ref{sec:exp} showing that the MMM introduced in the previous section is crucial to ensure that the pre-trained hidden states (for video or text) are a good initialization for retrieval tasks.

\subsection{Action Segmentation}
Action segmentation is to assign each frame of a video with one of the pre-defined labels.
This is similar to the named entity recognition (NER) task in NLP but on video frames.
% This is a video-only task. 
We feed in VLM with the whole video, a dummy text token, and an isolated attention mask. 
Then we add a classification head (with the number of pre-defined labels) on top of the hidden states for each video token in the last layer of VLM. 

\subsection{Action Step Localization}
In action step localization, each video belongs to a task with multiple steps, where each step is described as a short text.
Then each frame of a video needs to be aligned with a step in text form.
% This is akin to sequence labeling in NLP but the labeling set is not fixed but associated with each video.
The challenge for applying BERT to action step localization is similar to text-video retrieval: video frames need to be aligned with textual steps in joint space and it is almost impossible for pairwise video/text matching because the number of frame/text pairs is large. 
% The hidden states for video/text are expected to be isolated.

Similar to the text-video retrieval model, we also apply isolated attention masks to video and text.
The major difference is that we pass video and text separately to BERT.
This is because the video can be several minutes long (more than 100 tokens)
%so we apply sliding windows in videos 
but the number of text labels for each video is fixed (e.g. under 10).
To keep the format of BERT being consistent for multi-modal inputs, we add a dummy text token for video forwarding and a dummy video token for text, respectively.
For a given frame(video token), we compute the distribution of that frame over textual steps via dot products and the softmax function. 

\subsection{Multiple-choice VideoQA}
Multiple-choice VideoQA~\cite{yu2018joint} aligns each video with one out of several candidate answers in the text.
The major difference between action step localization and multiple-choice VideoQA is that the video hidden state is not on frame-level but sequence-level.
We apply isolated attention masks to BERT and forward video and text answers (with dummy tokens), respectively.
Then the answer with the maximum similarity with the video is  reported.
During fine-tuning, we apply contrastive loss on video-text similarity to rank answers.

%\noindent \textbf{Video Captioning}
\subsection{Video Captioning}
Another big challenge of using a single encoder is how to apply generative tasks (such as video captioning) without pre-training an explicit decoder.
We observe that a transformer decoder \cite{vaswani2017attention} has the following major differences from an encoder:
(1) an auto-regressive loss that does not allow a text token to see future tokens;
%but all video (source in machine translation) tokens and text tokens;
(2) a prediction head to generate texts.
To resolve (1), one can easily fine-tune the text segment of VLM as auto-regressive loss by passing in shifted tokens and a lower-triangle attention mask to the text segment, as shown in Figure~\ref{fig:ft}. 
To resolve (2), inspired by \cite{rothe-etal-2020-leveraging,zhou2020unified} that uses BERT as a decoder, one can re-use language model heads as prediction heads for generation.
Note that this setting has less architecture design than a standard transformer decoder (e.g. no explicit self-attention on text or cross-attention on video).
The implicit text decoder inside BERT shares self-attention with the video encoder so to save the total number of parameters.

%\noindent \textbf{Classification/Regression}
%\subsection{Classification/Regression}
%Lastly, similar to existing studies, we use the \texttt{[CLS]} token to indicate the hidden states of the whole video/text sequence and apply a prediction head for classification.

\section{Experiment}
\label{sec:exp}
\subsection{Dataset}

\subsubsection{Pre-training}
We adopt the Howto100M dataset \cite{miech2019howto100m} for pre-training, which contains instructional videos originally from YouTube via searching keywords from wikihow (\url{www.wikihow.com}). 
After filtering the unavailable ones, we get 1.1M videos. We split 4000 videos as the validation set and the rest for pre-training.
On average, the duration of each video is about 6.5 minutes with 110 clip-text pairs.
After removing repeated texts within overlapped clips from ASR, we get about 7.7+ GB texts of captions, with 2.4 tokens per second on average.
%Then, we fine-tune our pre-trained model on five diverse downstream tasks using five datasets, including text-based video retrieval, multimodal video captioning, action segmentation, action step localization, and multimodal sentiment analysis.

\subsubsection{Fine-tuning}
\noindent \textbf{MSR-VTT} \cite{xu2016msr} is a popular dataset for \textit{text-video retrieval} and \textit{VideoQA}.
It has open domain video clips, and each training clip has 20 captioning sentences labeled by humans. 
There are 200K clip-text pairs from 10K videos in 20 categories, including sports, music, etc.
Following JSFusion\cite{yu2018joint,miech2019howto100m}, we randomly sampled 1,000 clip-text pairs as test data.
%to evaluate on the text-based video retrieval task.
We further use the QA test data \cite{yu2018joint} as the dataset for multiple-choice VideoQA.

\noindent \textbf{Youcook2} \cite{zhou2017towards} contains 2,000 cooking videos on 89 recipes with 14K video clips from YouTube. 
The overall duration is 176 hours (5.26 minutes on average). 
Each video clip is annotated with one captioning sentence. 
Follow the split setting in\cite{miech2019howto100m}, we evaluate both text-based video retrieval and multimodal video captioning tasks. 
We filter the data and make sure there is no overlap between pre-training and evaluation data. 
After filtering out unavailable ones, we have 9,473 training clip-text pairs from 1222 videos and 3,305 test clip-text pairs from 430 videos.
%For the video captioning task, we use the same setting as in \cite{shi-etal-2019-dense}. 

\noindent \textbf{COIN} \cite{tang2019coin} are leveraged to evaluate \textit{action segmentation}.
It has 11,827 videos (476 hours) and each video is labeled with 3.91 step segments on average and 46,354 segments in total. There are 778 step labels, plus one background (\underline{O}utside) label. Since one video can last for several minutes that are much longer than the maximum length of the video segment of VLM. We apply a sliding window with step size 16 and window size 32.
During inference, we average the logits for overlapped frames from multiple windows.

\noindent \textbf{CrossTask} \cite{zhukov2019cross} is a dataset for action localization that contains 83 different tasks and 4.7k videos. Each task has a set of steps with text descriptions annotated on temporal frames of the video.
We use the testing data split via the official code\footnote{ \url{https://github.com/DmZhukov/CrossTask}}, which contains annotated 1690 videos. The rest of the 540 annotated videos are used for weakly supervised training.

%\noindent \textbf{MOSI} is a dataset for multimodal sentiment classification/regression \todo{WIP}.
\begin{table*}[t]
    \centering
    \scalebox{0.61}{
        \begin{tabular}{l c c c r c c}
            \hline
            Model & Paradigm & \#params. & \#loss & \#unimodal/cross en/decoder & Joint Retrieval & Generation\\
            \hline
            MMT\cite{gabeur2020multi} & task-specific alignment & 127.3M & 1 & 2/0/0 & yes & no\\
            ActBERT\cite{zhu2020actbert} & weakly supervised/MTL & n/a (3 typed attentions) & 4 & 0/1(modal-typed attn.)/0 & no(pair) & extra decoder \\
            VideoAsMT\cite{korbar2020video} & weakly supervised/MTL & 286M(base)/801M(large) & 1 & 1/1/1 & no (gen.) & yes\\
            HERO\cite{li-etal-2020-hero} & SSL(w/ sup. video feat.)/MTL & 159M & 5 & 1(query)/2/0 & no(pair) & extra decoder\\
            UniVL\cite{luo2020univilm} & SSL/MTL & 260M & 5 & 2/1/1 & yes & yes\\
            \hline
            VLM & SSL/Task-agnostic & \textbf{110M} & \textbf{1} & 0/\textbf{1}/0(shared w/ encoder) & yes & yes\\
            \hline
        \end{tabular}
        \vspace{-1mm}
    }
    \caption{Comparison of pre-trained models on learning paradigms (SSL means self-supervised learning; MTL means multi-task learning), number of parameters (\# params.), number of losses (\#loss), number of unimodal/cross-modal encoders/decoders, and whether to support retrieval in joint space(joint retrieval) and text generation. Types and numbers are estimated based on released code or papers: exceptions are in parenthesis (e.g. pair means pairwise matching using \texttt{[CLS]}). VLM is extremely simple with fewer parameters and limitations.}
    \label{tbl:num_param}
\end{table*}

\subsection{Hyper-parameters}
We extract video tokens from video frames using the S3D encoder pre-trained from \cite{miech2020end}. 
The fps is 30 and we extract one (1) video token per second with the dimension of 512. We apply an MLP to transform such 512 dimensions to the hidden size (768) of $\text{BERT}_{\text{BASE}}$.

Following \cite{luo2020univilm}, we adopt $\text{BERT}_{\text{BASE}}$ (uncased) as our base model and tuned directly from BERT's weights, so all hyper-parameters are the same as the original BERT. The maximum length of BERT is set as 96, where 32 tokens are for videos and the rest tokens are for text and special tokens.
Remind that texts are 2.4 tokens per second and video tokens are 1 token per second.
We form a text clip with a random length in-between 8 and 64 text tokens and collect the corresponding video clip to form a training example.
We randomly sample 32 video/text clip pairs from each video and use 8 videos to form a batch of size 256.
Each training example has 50\% chance for MMM (25\% for whole video masking and 25\% for whole text masking) and 50\% chance on MFM-MLM (with 15\% probability of video and text token masking).

We pre-train VLM on 8 NVIDIA Tesla V100 GPUs (each with 32 GB memory) for 15 epochs using fp16 for one (1) day.
Following \cite{liu2019roberta}, we choose Adam \cite{kingma2014adam} optimizer with initial learning rate of 5e-5 (with betas as (0.9, 0.98)), 1000 steps of warm-up and a polynomial decay learning rate scheduler. Gradients are clipped with 2.0. All fine-tuning tasks use the same hyper-parameters as pre-training except the number of warm-up steps is 122.

\subsection{Model Comparison}
We first investigate the design choices of VLM compared to other transformer-based multimodal pre-training baselines.
As shown in Table~\ref{tbl:num_param}, we collect training paradigms, model sizes, etc. of these models (estimated based on their source codes or papers).
VLM is significantly smaller than other models since it is just a $\text{BERT}_{\text{BASE}}$ (uncased), while it is still fully self-supervised, task-agnostic (e.g. no training on retrieval or auto-regressive style tasks) and supports joint retrieval and text generation.
% With 7.7+ GB of text captions in Howto100M, we suspect existing models are over-parameterized since $\text{BERT}_{\text{BASE}}$ typically can consume 10GB+ corpora for LMs.

\subsection{Quantitative Analysis}
We investigate the performance of VLM on fine-tuning tasks with very basic setups (e.g. no augmented features, large LMs, optimized losses for particular tasks).
Note that it could be hard for fair comparisons between task-agnostic and task-specific approaches. We list other baselines by type and our goal is a simple baseline for task-agnostic pre-training as better initialization of strongly performed fine-tuning models.

\noindent \textbf{Text-video Retrieval} We use MSR-VTT and Youcook2 to evaluate the performance on text-video retrieval.
The results are shown in Table~\ref{tbl:vtt} and \ref{tbl:youcook}, respectively.
VLM achieves good performance on these two datasets, indicating that the MMM and isolated self-attention mask can be used together for joint retrieval. Ablation study shows that using an isolated self-attention mask alone does not yield good performance, indicating MMM is very important to learn features for alignment. Note that our pre-training is task-agnostic but still outperforms baselines with retrieval style pre-training.

\begin{table}[t]
\centering
\scalebox{0.63}{
    \begin{tabular}{l c c c c}
    \hline
    Methods & R@1 & R@5 & R@10 & Median R \\
    \hline
    Random & 0.1 & 0.5 & 1.0 & 500 \\
    % C+LSTM+SA\cite{torabi2016learning} & 4.2 & 12.9 & 19.9 & 55 \\
%     VSE\cite{kiros2014unifying} & 3.8 & 12.7 & 17.1 & 66 \\
    % SNUVL\cite{yu2016video} & 3.5 & 15.9 & 23.8 & 44 \\
    % \cite{kaufman2017temporal} & 4.7 & 16.6 & 24.1 & 41 \\
    % CT-SAN\cite{yu2017end} & 4.4 & 16.6 & 22.3 & 35 \\
    % JSFusion\cite{yu2018joint} & 10.2 & 31.2 & 43.2 & 13 \\
%    HowTo100M\cite{miech2019howto100m} & 14.9 & 40.2 & 52.8 & 9 \\
%   MIL-NCE\cite{miech2020end} & 9.9 & 24.0 & 32.4 & 29.5 \\
    \hline
    Task-specific Alignment Pre-training\\
    MMT \cite{gabeur2020multi} & 25.8 & 57.2 & 69.3 & 4\\
    \hline
    Pairwise Matching\\
    ActBERT\cite{zhu2020actbert} & 8.6 & 23.4 & 33.1 & 36 \\
    VideoAsMT\cite{korbar2020video} & 14.7 & - & 52.8 & - \\
    \hline
    Multi-task Pre-training\\
    HERO \cite{li-etal-2020-hero} &  16.80 & 43.40 & 57.70 & - \\
    UniVL (FT-Joint) \cite{luo2020univilm} & 20.6 & 49.1 & 62.9 & 6 \\
    \hline
    %UniVL (FT-Align)\cite{luo2020univilm} & 21.2 & 49.6 & 63.1 & 6 \\
    VLM & 28.10 & 55.50 & 67.40 & 4\\
    % VLM & 25.00 & 51.80 & 64.60 & 5.0\\
    % VLM (MFM+MLM)\todo{outofdate} & 23.8 & 53.0 & 66.4 & 5 \\
%     Ours (Modal Gen) & 22.5 & 52.0 & 63.8 & 5 \\
    \hline
    \end{tabular}
}
\caption{Results of text-video retrieval on MSR-VTT dataset.}
\label{tbl:vtt}
% \vspace{-4mm}
\end{table}

\begin{table}[t]
\centering
\scalebox{0.63}{
    \begin{tabular}{l c c c c}
    \hline
    Methods & R@1 & R@5 & R@10 & Median R \\
    \hline
    Random & 0.03 & 0.15 & 0.3  & 1675 \\
    % HGLMM\cite{klein2015associating} & 4.6 & 14.3 & 21.6  & 75 \\
    % HowTo100M\cite{miech2019howto100m} & 8.2 & 24.5 & 35.3  & 24 \\
    % MIL-NCE\cite{miech2020end} & 15.1 & 38.0 & 51.2  & 10 \\
    \hline
    Task-specific Alignment Pre-training\\
    Coot\cite{ging2020coot} & 16.7 & 40.2 & 52.3 & 9\\
    \hline
    Pairwise Matching\\
    ActBERT\cite{zhu2020actbert} & 9.6 & 26.7 & 38.0  & 19 \\
    VideoAsMT\cite{korbar2020video} & 11.6 & - & 43.9 &  - \\
    \hline
    Multi-task Pre-training\\
    UniVL (FT-Joint)\cite{luo2020univilm} & 22.2 & 52.2 & 66.2 & 5 \\
    % UniVL (FT-Align)\cite{luo2020univilm} & 28.9 & 57.6 & 70.0 & 4 \\
    \hline
    VLM & 27.05 & 56.88 & 69.38 & 4\\
    % VLM (MFM+MLM)\todo{outofdate} & 25.72 & 54.74 & 69.35 & 5 \\
    \hline
    \end{tabular}
}
\caption{Results of text-based video retrieval on Youcook2 dataset.}
\label{tbl:youcook}
% \vspace{-4mm}
\end{table}

\noindent \textbf{Action Segmentation} We report the results of action segmentation on COIN dataset in Table~\ref{tbl:coin}. VLM outperforms other baselines indicating its good token-level video representations. Note that this task only tests the hidden states of the video indicating the unimodal encoding capability of VLM is not compromised.

\begin{table}[t]
\centering
\scalebox{0.8}{
    \begin{tabular}{l c}
    \hline
Method & Frame Accuracy \\
\hline
NN-Viterbi \cite{richard2018neuralnetwork} & 21.17\\
VGG \cite{simonyan2014very} & 25.79\\
TCFPN-ISBA \cite{ding2018weakly} & 34.30\\
CBT \cite{sun2019contrastive} & 53.90\\
MIL-NCE \cite{miech2020end} & 61.00\\
ActBERT \cite{zhu2020actbert} & 56.95\\
\hline
VLM & 68.39 \\
    \hline
    \end{tabular}
}
\caption{Action segmentation on COIN dataset.}
% (We didn't report the score from UniVL as it uses task classes instead of step labels)
\label{tbl:coin}
% \vspace{-4mm}
\end{table}

\noindent \textbf{Action Step Localization} We setup two (2) evaluations for the CrossTask dataset.
First, we evaluate the zero-shot transfer of VLM. Note that existing studies evaluate Crosstask with retrieval/alignment style pre-training, where the aligned hidden states are directly used for action step localization.
Our task-agnostic pre-training derives an even harder problem: applying hidden states learned from proxy tasks on video frame/text alignment for action step localization without explicitly training on alignment.
We simply use the hidden states from the last layer of VLM for video/text representation and directly compute the similarities between video frames and text descriptions.
Surprisingly, the performance is better than some baselines and closer to one supervised method. This indicates masked token loss together with MMM can learn certain video-text alignments in joint space.
Second, we use just 540 videos for weakly supervised training and we get a much better result.

\begin{table}
\centering
\scalebox{0.82}{
    \begin{tabular}{l c}
\hline
Methods & Average Recall\\
\hline
Joint Alignment\\
Alayrac \cite{alayrac2016unsupervised} & 13.3\\
Zhukov \cite{zhukov2019cross} & 22.4\\
Supervised \cite{zhukov2019cross} & 31.6\\
HowTo100M \cite{miech2019howto100m} & 33.6\\
MIL-NCE \cite{miech2020end} & 40.5\\
UniVL \cite{luo2020univilm} & 42.0\\
\hline
Pairwise Matching\\
ActBERT \cite{zhu2020actbert} & 41.4\\
\hline
VLM (task-agnostic, zero-shot) & 28.5 \\
VLM (supervised on 540 videos) & 46.5 \\
    \hline
    \end{tabular}
}
\caption{Action step localization results on CrossTask.}
% \vspace{-4mm}
\end{table}

\noindent \textbf{Video Question Answering} We use MSR-VTT QA 
%\todo{there are two versions of this, probably we follow JSFusion/ActBERT} 
to evaluate multiple-choice question answering.
Recall that this task essentially tests video-text similarity.
The performance of VLM is better than ActBERT, which leverages pairwise matching for each video/answer pair.
% This indicates that VLM has a strong capability of video/text alignment after fine-tuning.
%Similar to text-video retrieval.
%We feed video-text pairs into VLM to learn a retrieval model and use the pooled latent space to compute the dot product of a pair.
%At the inference time, we fed each candidate
%with the video clip to VLM. The final choice is made
%by selecting the candidates with the max similarity score.

\begin{table}
\centering
\scalebox{0.8}{
    \begin{tabular}{l c}
    \hline
Method & Accuracy \\
\hline
% LSTM-fusion \cite{yu2018joint} & 38.3 \\
% C+LSTM+SA-FC7 \cite{torabi2016learning} & 60.2 \\
% VSE-LSTM \cite{kiros2014unifying} & 67.3 \\
% SNUVL \cite{yu2016video} & 65.4 \\
% EITanque \cite{kaufman2017temporal} & 65.5 \\
% CT-SAN \cite{yu2017end} & 66.4 \\
% MLB \cite{kim2016hadamard} & 76.1 \\
Joint Retrieval\\
JSFusion\cite{yu2018joint} & 83.4 \\
\hline
Pairwise Matching\\
ActBERT\cite{zhu2020actbert} & 85.7\\
\hline
VLM & 91.64 \\
    \hline
    \end{tabular}
}
\caption{Video question answering (multiple-choices) evaluated on MSR-VTT.}
% \vspace{-4mm}
\end{table}

\noindent \textbf{Video Captioning} We lastly evaluate VLM on video captioning with autoregressive attention mask with other baselines that have an explicit text decoder.
As shown in Table~\ref{tbl:youcookcap}, our ``compact'' decoder using BERT's LM heads is surprisingly good at video captioning compared to other fine-tuning baselines with external decoders (e.g. Coot).
This indicates that it is possible to remove an explicit decoder and sharing weights between video and text tokens.
% Later study shows that MMM is very important in fusing video/text features within a single encoder.

\begin{table}[t]
\centering
\scalebox{0.62}{
    \begin{tabular}{l c c c c c}
    \hline
    Methods & B-3 & B-4 & M & R-L & CIDEr \\
    \hline
% Bi-LSTM \cite{zhou2017towards} & - & 0.87 & 8.15 & - & - \\
% EMT \cite{zhou2018end} & - & 4.38 & 11.55 & 27.44 & 0.38 \\
Extra Decoder\\
VideoBERT \cite{sun2019videobert} & 6.80 & 4.04 & 11.01 & 27.50 & 0.49 \\
CBT \cite{sun2019contrastive} & - & 5.12 & 12.97 & 30.44 & 0.64 \\
ActBERT \cite{zhu2020actbert} & 8.66 & 5.41 & 13.30 & 30.56 & 0.65 \\
Coot\cite{ging2020coot} & 17.62 & 11.09 & 19.34 & 37.63 & -\\
\hline
w/ Pre-trained Decoder \\
VideoAsMT \cite{korbar2020video} & - & 5.3 & 13.4 & - & - \\
% AT \cite{hessel2019case} & T & - & 8.55 & 16.93 & 35.54 & 1.06 \\
% DPC \cite{shi-etal-2019-dense} & V + T & 7.60 & 2.76 & 18.08 & - & - \\
% AT+Video \cite{hessel2019case} & V + T & - & 9.01 & 17.77 & 36.65 & 1.12 \\
UniVL \cite{luo2020univilm} & 16.46 & 11.17 & 17.57 & 40.09 & 1.27 \\
% UniVL \cite{luo2020univilm} & T & 20.32 & 14.70 & 19.39 & 41.10 & 1.51 \\
% UniVL \cite{luo2020univilm} & V + T & 23.87 & 17.35 & 22.35 & 46.52 & 1.81 \\
\hline
VLM & 17.78 & 12.27 & 18.22 & 41.51 & 1.3869\\
% VLM (MFM+MLM)\todo{outofdate} & V & 17.26 & 11.85 & 17.79 & 40.69 & 1.3228\\
% Ours (Modal Gen) & V & 17.50 & 12.04 & 18.20 & 40.66 & 1.3464\\
    \hline
    \end{tabular}
}
\caption{Video captioning results on Youcook2 dataset.}
\label{tbl:youcookcap}
% \vspace{-4mm}
\end{table}

\subsubsection{Ablation Study}
We use Youcook2 as the base task for the ablation study on text-retrieval and video captioning. 
We are interested in the following study:
(1) percentage of examples for MMM (w/ MMM x\%);
(2) minimum length of text tokens, where the length of video will be determined by the start/end timestamps of text tokens;
(3) performance of $\mathcal{L}_\text{VLM}$ (Equation \ref{eq:lvlm}).
The results are shown in Table~\ref{tbl:abl_retri} and Table~\ref{tbl:abl_cap}.

\noindent \textbf{Effects of MMM} Without MMM (w/ MMM 0\%, or MFM-MLM 100\%), the performance significantly dropped. This indicates that a naive adoption of traditional MFM-MLM masking may not learn joint video/text representations well, as indicated by both retrieval and captioning task. We suspect a masked token is more likely predicted from tokens of the same modality.
We further try MMM with different probabilities (30\% or 70\%) and 50\% is the best.

\noindent \textbf{Minimum Length of Texts} The length of a clip can be important for retrieval tasks \cite{miech2020end}. We ran VLM on longer (at least 16 text tokens) video/text pairs. The performance is slightly dropped, indicating pre-training on longer clips may not cover fine-tuning tasks with short clips.

\noindent \textbf{Effects of Masked Token Loss} We notice that using multi-task style loss $\mathcal{L}_{\text{MFM-MLM}}$ may reduce the performance. This indicates learning a masked token from both video/text tokens can help.

\begin{table}[!htbp]
\centering
\scalebox{0.78}{
    \begin{tabular}{l c c c c}
    \hline
    VLM & R@1 & R@5 & R@10 & Median R \\
    \hline
    w/ MMM 50\% & 27.05 & 56.88 & 69.38 & 4.0\\
    w/ MMM 0\% & 15.12 & 39.47 & 52.81 & 9.0\\
    w/ MMM 30\% & 25.30 & 54.80 & 68.96 & 4.0\\
    w/ MMM 70\% & 25.17 & 54.98 & 69.11 & 4.0\\
    w/ min. 16 text tokens & 25.84 & 54.43 & 68.29 & 5.0\\
    w/ $\mathcal{L}_{\text{MFM-MLM}}$ & 26.93 & 55.92 & 69.86 & 4.0\\
    \hline
    \end{tabular}
}

% MTM & 25.84 & 54.43 & 68.29 & 5.0\\
% MFM/MLM & 24.36 & 53.62 & 67.17 & 5.0\\
% MTM - MMM & 15.37 & 39.94 & 54.10 & 9.0\\
% MFM/MLM - MMM & 14.64 & 39.46 & 53.31 & 9.0\\

\caption{Ablation study of VLM for text-based video retrieval on Youcook2.}
\label{tbl:abl_retri}
\end{table}

\begin{table}[!htbp]
\centering
\scalebox{0.72}{
    \begin{tabular}{l c c c c c}
    \hline
    VLM & B-3 & B-4 & M & R-L & CIDEr \\
    \hline
    w/ MMM 50\% & 17.78 & 12.27 & 18.22 & 41.51 & 1.3869\\
    w/ MMM 0\% & 15.47 & 10.54 & 16.49 & 38.83 & 1.2163\\
    w/ MMM 30\% & 16.57 & 11.30 & 17.55 & 40.76 & 1.3215\\
    w/ MMM 70\% & 16.94 & 11.68 & 17.67 & 41.24 & 1.3739\\
    w/ min. 16 text tokens & 17.25 & 12.00 & 17.67 & 40.62 & 1.3076 \\
    w/ $\mathcal{L}_{\text{MFM-MLM}}$ & 16.66 & 11.53 & 17.34 & 40.36 & 1.3224\\
    \hline
    \end{tabular}
}

% MTM & 17.25 & 12.00 & 17.67 & 40.62 & 1.3076 \\
% MFM/MLM & 16.25 & 11.20 & 17.26 & 40.55 & 1.3241\\
% MTM - MMM & 15.84 & 10.79 & 16.66 & 39.20 & 1.2354\\
% MFM/MLM - MMM & 15.10 & 10.32 & 16.05 & 38.49 & 1.2079\\

\caption{Ablation study of VLM for video captioning on Youcook2 dataset.}
\label{tbl:abl_cap}
\end{table}

% \subsubsection{Video Encoders}
% We further explore different video encoders and their effects on performance.

% \begin{table}[!htbp]
% \centering
% \scalebox{0.6}{
%     \begin{tabular}{l|c|c|c|c}
%     \hline
%     Video Encoder & R@1 & R@5 & R@10 & Median R \\
%     \hline
%     S3D & 25.84 & 54.43 & 68.29 & 5.0\\
%     SlowFast\cite{feichtenhofer2019slowfast} & & & \\
%     S3D (sup. on Kinetics) & & & \\
%     \hline
%     \end{tabular}
% }
% \caption{Ablation study of VLM for text-based video retrieval on Youcook2.}
% \label{tbl:youcook}
% \end{table}

\subsection{Qualitative Analysis}
\subsubsection{Error Analysis} 
\noindent \textbf{Text-video retrieval}.
We use MSR-VTT as the dataset for error analysis on text-video retrieval, as shown in Table~\ref{tbl:err_vtt} of Appendix.
We pair the query text with the text of the top-1 ranked video to show 100 errors in ranking since video tokens are harder to present.
We observe the following types of errors in video understanding: (1) objects sometimes are hard to recognize such as dog or cat; (2) attributes of objects may be hard to match the text, e.g. gender, ages, etc. (3) subtle differences of actions;  (4) specific videos for a general query or vice versa, e.g. people vs basketball player. We believe the last type may not be errors but hard for existing annotations or evaluations to separate.

\noindent \textbf{Video Captioning}.
We further examine the generated text from video captioning.
Note that our video captioning has no support from ASR or transcript so the video is the only source to generate text content and errors of video understanding can easily be reflected in the text.
From Table~\ref{tbl:error_youcookcap} of Appendix, we notice that one major type of error is from objects of similar shapes and colors, e.g. onion rings vs shrimp.

\subsubsection{Visualization}.
% We first examine the video/text tokens in the joint embedding space (before the summation with positional embeddings), as in Figure~\ref{fig:token_embedding} in Appendix.
We observe that video tokens take the majority of space while text tokens are rather clustered together. This is probably because videos from the physical world are more diverse and sparse than text from a fixed vocabulary.

We plot the self-attention of VLM layers within and in-between each modality, as in Figure~\ref{fig:attn0_1} of Appendix.
We observe the following patterns from all 144 attention heads:
\begin{itemize}
    \item Unlike LMs, there are no recurrent (shifted) position-wise patterns for video tokens;
    \item Self-attentions in the 1st layer are more diverse than later layers. This suggests that existing video encoders might be too deep for transformers;
    \item Some attention heads show patterns of cross-modal mapping in-between video and text (e.g. sub-figure (a));
    \item Word-level cross-modal co-reference: video tokens with \textit{pouring soy sauce} refers to the text token of ``soy'' (e.g. sub-figure (b));
\end{itemize}

\section{Conclusions}
%This paper proposes a simple baseline that leverages the original BERT for multimodal pre-training.
%We propose to learn a fused feature space end-to-end on BERT transformers on vision and text.
%We further proposes unify masked frame model and masked language model as masked token modal and propose masked modality model to encourage learning modality-level features for retrieval and generation.
We presented a task-agnostic pre-training with new masking schemes that enable the training of a single masked language model that can accept either video or text input, or both.
We showed that this simple VLM model can be effectively tuned for a broad range of downstream tasks, such as text-video retrieval and video captioning via different types of attention masks.
% \todo{We further augment pre-training with a retrieval model built on top of the hidden states of BERT.[decide later whether to keep the previous argument or retrieval model alone.]}
Experimental results show that the proposed methods maintain competitive performance while requiring a significantly smaller number of parameters than competing methods.

\section*{Acknowledgments}
We thank Huaishao Luo (author of UniVL\cite{luo2020univilm}), Mandela Patrick (author of Support-Set\cite{patrick2021supportset}) and Luowei Zhou (author of Youcook\cite{zhou2017towards}) for supports of baseline setup.

\bibliographystyle{acl_natbib}
\bibliography{acl2021}

\begin{thebibliography}{40}
\expandafter\ifx\csname natexlab\endcsname\relax\def\natexlab#1{#1}\fi

\bibitem[{Alayrac et~al.(2016)Alayrac, Bojanowski, Agrawal, Sivic, Laptev, and
  Lacoste-Julien}]{alayrac2016unsupervised}
Jean-Baptiste Alayrac, Piotr Bojanowski, Nishant Agrawal, Josef Sivic, Ivan
  Laptev, and Simon Lacoste-Julien. 2016.
\newblock Unsupervised learning from narrated instruction videos.
\newblock In \emph{Proceedings of the IEEE Conference on Computer Vision and
  Pattern Recognition}, pages 4575--4583.

\bibitem[{Alayrac et~al.(2020)Alayrac, Recasens, Schneider, Arandjelovi{\'c},
  Ramapuram, De~Fauw, Smaira, Dieleman, and Zisserman}]{alayrac2020self}
Jean-Baptiste Alayrac, Adri{\`a} Recasens, Rosalia Schneider, Relja
  Arandjelovi{\'c}, Jason Ramapuram, Jeffrey De~Fauw, Lucas Smaira, Sander
  Dieleman, and Andrew Zisserman. 2020.
\newblock Self-supervised multimodal versatile networks.
\newblock \emph{arXiv preprint arXiv:2006.16228}.

\bibitem[{Chen et~al.(2020)Chen, Li, Yu, El~Kholy, Ahmed, Gan, Cheng, and
  Liu}]{chen2020uniter}
Yen-Chun Chen, Linjie Li, Licheng Yu, Ahmed El~Kholy, Faisal Ahmed, Zhe Gan,
  Yu~Cheng, and Jingjing Liu. 2020.
\newblock Uniter: Universal image-text representation learning.
\newblock In \emph{European Conference on Computer Vision}, pages 104--120.
  Springer.

\bibitem[{Clark et~al.(2019)Clark, Khandelwal, Levy, and
  Manning}]{clark2019does}
Kevin Clark, Urvashi Khandelwal, Omer Levy, and Christopher~D Manning. 2019.
\newblock What does bert look at? an analysis of bert’s attention.
\newblock In \emph{Proceedings of the 2019 ACL Workshop BlackboxNLP: Analyzing
  and Interpreting Neural Networks for NLP}, pages 276--286.

\bibitem[{Devlin et~al.(2019)Devlin, Chang, Lee, and
  Toutanova}]{devlin-etal-2019-bert}
Jacob Devlin, Ming-Wei Chang, Kenton Lee, and Kristina Toutanova. 2019.
\newblock \href {https://doi.org/10.18653/v1/N19-1423} {{BERT}: Pre-training of
  deep bidirectional transformers for language understanding}.
\newblock In \emph{Proceedings of the 2019 Conference of the North {A}merican
  Chapter of the Association for Computational Linguistics: Human Language
  Technologies, Volume 1 (Long and Short Papers)}, pages 4171--4186,
  Minneapolis, Minnesota. Association for Computational Linguistics.

\bibitem[{Ding and Xu(2018)}]{ding2018weakly}
Li~Ding and Chenliang Xu. 2018.
\newblock Weakly-supervised action segmentation with iterative soft boundary
  assignment.
\newblock In \emph{Proceedings of the IEEE Conference on Computer Vision and
  Pattern Recognition}, pages 6508--6516.

\bibitem[{Gabeur et~al.(2020)Gabeur, Sun, Alahari, and
  Schmid}]{gabeur2020multi}
Valentin Gabeur, Chen Sun, Karteek Alahari, and Cordelia Schmid. 2020.
\newblock Multi-modal transformer for video retrieval.
\newblock In \emph{European Conference on Computer Vision (ECCV)}, volume~5.
  Springer.

\bibitem[{Ging et~al.(2020)Ging, Zolfaghari, Pirsiavash, and
  Brox}]{ging2020coot}
Simon Ging, Mohammadreza Zolfaghari, Hamed Pirsiavash, and Thomas Brox. 2020.
\newblock Coot: Cooperative hierarchical transformer for video-text
  representation learning.
\newblock \emph{arXiv preprint arXiv:2011.00597}.

\bibitem[{Huang et~al.(2021)Huang, Patrick, Hu, Neubig, Metze, and
  Hauptmann}]{multiht100m_bernie}
Po{-}Yao Huang, Mandela Patrick, Junjie Hu, Graham Neubig, Florian Metze, and
  Alexander~G. Hauptmann. 2021.
\newblock \href {http://arxiv.org/abs/2103.08849} {Multilingual multimodal
  pre-training for zero-shot cross-lingual transfer of vision-language models}.
\newblock \emph{CoRR}, abs/2103.08849.

\bibitem[{Kingma and Ba(2014)}]{kingma2014adam}
Diederik~P Kingma and Jimmy Ba. 2014.
\newblock Adam: A method for stochastic optimization.
\newblock \emph{arXiv preprint arXiv:1412.6980}.

\bibitem[{Korbar et~al.(2020)Korbar, Petroni, Girdhar, and
  Torresani}]{korbar2020video}
Bruno Korbar, Fabio Petroni, Rohit Girdhar, and Lorenzo Torresani. 2020.
\newblock Video understanding as machine translation.
\newblock \emph{arXiv preprint arXiv:2006.07203}.

\bibitem[{Lewis et~al.(2020)Lewis, Liu, Goyal, Ghazvininejad, Mohamed, Levy,
  Stoyanov, and Zettlemoyer}]{lewis-etal-2020-bart}
Mike Lewis, Yinhan Liu, Naman Goyal, Marjan Ghazvininejad, Abdelrahman Mohamed,
  Omer Levy, Veselin Stoyanov, and Luke Zettlemoyer. 2020.
\newblock \href {https://doi.org/10.18653/v1/2020.acl-main.703} {{BART}:
  Denoising sequence-to-sequence pre-training for natural language generation,
  translation, and comprehension}.
\newblock In \emph{Proceedings of the 58th Annual Meeting of the Association
  for Computational Linguistics}, pages 7871--7880, Online. Association for
  Computational Linguistics.

\bibitem[{Li et~al.(2020{\natexlab{a}})Li, Duan, Fang, Gong, Jiang, and
  Zhou}]{li2020unicoder}
Gen Li, Nan Duan, Yuejian Fang, Ming Gong, Daxin Jiang, and Ming Zhou.
  2020{\natexlab{a}}.
\newblock Unicoder-vl: A universal encoder for vision and language by
  cross-modal pre-training.
\newblock In \emph{AAAI}, pages 11336--11344.

\bibitem[{Li et~al.(2020{\natexlab{b}})Li, Chen, Cheng, Gan, Yu, and
  Liu}]{li-etal-2020-hero}
Linjie Li, Yen-Chun Chen, Yu~Cheng, Zhe Gan, Licheng Yu, and Jingjing Liu.
  2020{\natexlab{b}}.
\newblock \href {https://doi.org/10.18653/v1/2020.emnlp-main.161} {{HERO}:
  Hierarchical encoder for {V}ideo+{L}anguage omni-representation
  pre-training}.
\newblock In \emph{Proceedings of the 2020 Conference on Empirical Methods in
  Natural Language Processing (EMNLP)}, pages 2046--2065, Online. Association
  for Computational Linguistics.

\bibitem[{Li et~al.(2019)Li, Yatskar, Yin, Hsieh, and Chang}]{li2019visualbert}
Liunian~Harold Li, Mark Yatskar, Da~Yin, Cho-Jui Hsieh, and Kai-Wei Chang.
  2019.
\newblock Visualbert: A simple and performant baseline for vision and language.
\newblock In \emph{Arxiv}.

\bibitem[{Liu et~al.(2019)Liu, Ott, Goyal, Du, Joshi, Chen, Levy, Lewis,
  Zettlemoyer, and Stoyanov}]{liu2019roberta}
Yinhan Liu, Myle Ott, Naman Goyal, Jingfei Du, Mandar Joshi, Danqi Chen, Omer
  Levy, Mike Lewis, Luke Zettlemoyer, and Veselin Stoyanov. 2019.
\newblock Roberta: A robustly optimized bert pretraining approach.
\newblock \emph{arXiv preprint arXiv:1907.11692}.

\bibitem[{Lu et~al.(2019)Lu, Batra, Parikh, and Lee}]{lu2019vilbert}
Jiasen Lu, Dhruv Batra, Devi Parikh, and Stefan Lee. 2019.
\newblock \href
  {https://proceedings.neurips.cc/paper/2019/file/c74d97b01eae257e44aa9d5bade97baf-Paper.pdf}
  {Vilbert: Pretraining task-agnostic visiolinguistic representations for
  vision-and-language tasks}.
\newblock In \emph{Advances in Neural Information Processing Systems},
  volume~32, pages 13--23. Curran Associates, Inc.

\bibitem[{Luo et~al.(2020)Luo, Ji, Shi, Huang, Duan, Li, Chen, and
  Zhou}]{luo2020univilm}
Huaishao Luo, Lei Ji, Botian Shi, Haoyang Huang, Nan Duan, Tianrui Li, Xilin
  Chen, and Ming Zhou. 2020.
\newblock Univilm: A unified video and language pre-training model for
  multimodal understanding and generation.
\newblock \emph{arXiv preprint arXiv:2002.06353}.

\bibitem[{Miech et~al.(2020)Miech, Alayrac, Smaira, Laptev, Sivic, and
  Zisserman}]{miech2020end}
Antoine Miech, Jean-Baptiste Alayrac, Lucas Smaira, Ivan Laptev, Josef Sivic,
  and Andrew Zisserman. 2020.
\newblock End-to-end learning of visual representations from uncurated
  instructional videos.
\newblock In \emph{Proceedings of the IEEE/CVF Conference on Computer Vision
  and Pattern Recognition}, pages 9879--9889.

\bibitem[{Miech et~al.(2019)Miech, Zhukov, Alayrac, Tapaswi, Laptev, and
  Sivic}]{miech2019howto100m}
Antoine Miech, Dimitri Zhukov, Jean-Baptiste Alayrac, Makarand Tapaswi, Ivan
  Laptev, and Josef Sivic. 2019.
\newblock Howto100m: Learning a text-video embedding by watching hundred
  million narrated video clips.
\newblock In \emph{Proceedings of the IEEE international conference on computer
  vision}, pages 2630--2640.

\bibitem[{Patrick et~al.(2021)Patrick, Huang, Asano, Metze, Hauptmann,
  Henriques, and Vedaldi}]{patrick2021supportset}
Mandela Patrick, Po-Yao Huang, Yuki Asano, Florian Metze, Alexander~G
  Hauptmann, Joao~F. Henriques, and Andrea Vedaldi. 2021.
\newblock \href {https://openreview.net/forum?id=EqoXe2zmhrh} {Support-set
  bottlenecks for video-text representation learning}.
\newblock In \emph{International Conference on Learning Representations}.

\bibitem[{Peters et~al.(2018)Peters, Neumann, Iyyer, Gardner, Clark, Lee, and
  Zettlemoyer}]{peters-etal-2018-deep}
Matthew Peters, Mark Neumann, Mohit Iyyer, Matt Gardner, Christopher Clark,
  Kenton Lee, and Luke Zettlemoyer. 2018.
\newblock \href {https://doi.org/10.18653/v1/N18-1202} {Deep contextualized
  word representations}.
\newblock In \emph{Proceedings of the 2018 Conference of the North {A}merican
  Chapter of the Association for Computational Linguistics: Human Language
  Technologies, Volume 1 (Long Papers)}, pages 2227--2237, New Orleans,
  Louisiana. Association for Computational Linguistics.

\bibitem[{Richard et~al.(2018)Richard, Kuehne, Iqbal, and
  Gall}]{richard2018neuralnetwork}
Alexander Richard, Hilde Kuehne, Ahsan Iqbal, and Juergen Gall. 2018.
\newblock Neuralnetwork-viterbi: A framework for weakly supervised video
  learning.
\newblock In \emph{Proceedings of the IEEE Conference on Computer Vision and
  Pattern Recognition}, pages 7386--7395.

\bibitem[{Rothe et~al.(2020)Rothe, Narayan, and
  Severyn}]{rothe-etal-2020-leveraging}
Sascha Rothe, Shashi Narayan, and Aliaksei Severyn. 2020.
\newblock \href {https://doi.org/10.1162/tacl_a_00313} {Leveraging pre-trained
  checkpoints for sequence generation tasks}.
\newblock \emph{Transactions of the Association for Computational Linguistics},
  8:264--280.

\bibitem[{Rouditchenko et~al.(2020)Rouditchenko, Boggust, Harwath, Joshi,
  Thomas, Audhkhasi, Feris, Kingsbury, Picheny, Torralba
  et~al.}]{rouditchenko2020avlnet}
Andrew Rouditchenko, Angie Boggust, David Harwath, Dhiraj Joshi, Samuel Thomas,
  Kartik Audhkhasi, Rogerio Feris, Brian Kingsbury, Michael Picheny, Antonio
  Torralba, et~al. 2020.
\newblock Avlnet: Learning audio-visual language representations from
  instructional videos.
\newblock \emph{arXiv preprint arXiv:2006.09199}.

\bibitem[{Simonyan and Zisserman(2014)}]{simonyan2014very}
Karen Simonyan and Andrew Zisserman. 2014.
\newblock Very deep convolutional networks for large-scale image recognition.
\newblock \emph{arXiv preprint arXiv:1409.1556}.

\bibitem[{Su et~al.(2020)Su, Zhu, Cao, Li, Lu, Wei, and Dai}]{Su2020VL-BERT:}
Weijie Su, Xizhou Zhu, Yue Cao, Bin Li, Lewei Lu, Furu Wei, and Jifeng Dai.
  2020.
\newblock \href {https://openreview.net/forum?id=SygXPaEYvH} {Vl-bert:
  Pre-training of generic visual-linguistic representations}.
\newblock In \emph{International Conference on Learning Representations}.

\bibitem[{Sun et~al.(2019{\natexlab{a}})Sun, Baradel, Murphy, and
  Schmid}]{sun2019contrastive}
Chen Sun, Fabien Baradel, Kevin Murphy, and Cordelia Schmid.
  2019{\natexlab{a}}.
\newblock Contrastive bidirectional transformer for temporal representation
  learning.
\newblock \emph{arXiv preprint arXiv:1906.05743}, 3(5).

\bibitem[{Sun et~al.(2019{\natexlab{b}})Sun, Myers, Vondrick, Murphy, and
  Schmid}]{sun2019videobert}
Chen Sun, Austin Myers, Carl Vondrick, Kevin Murphy, and Cordelia Schmid.
  2019{\natexlab{b}}.
\newblock Videobert: A joint model for video and language representation
  learning.
\newblock In \emph{Proceedings of the IEEE International Conference on Computer
  Vision}, pages 7464--7473.

\bibitem[{Tan and Bansal(2019)}]{tan-bansal-2019-lxmert}
Hao Tan and Mohit Bansal. 2019.
\newblock \href {https://doi.org/10.18653/v1/D19-1514} {{LXMERT}: Learning
  cross-modality encoder representations from transformers}.
\newblock In \emph{Proceedings of the 2019 Conference on Empirical Methods in
  Natural Language Processing and the 9th International Joint Conference on
  Natural Language Processing (EMNLP-IJCNLP)}, pages 5100--5111, Hong Kong,
  China. Association for Computational Linguistics.

\bibitem[{Tang et~al.(2019)Tang, Ding, Rao, Zheng, Zhang, Zhao, Lu, and
  Zhou}]{tang2019coin}
Yansong Tang, Dajun Ding, Yongming Rao, Yu~Zheng, Danyang Zhang, Lili Zhao,
  Jiwen Lu, and Jie Zhou. 2019.
\newblock Coin: A large-scale dataset for comprehensive instructional video
  analysis.
\newblock In \emph{Proceedings of the IEEE Conference on Computer Vision and
  Pattern Recognition}, pages 1207--1216.

\bibitem[{Vaswani et~al.(2017)Vaswani, Shazeer, Parmar, Uszkoreit, Jones,
  Gomez, Kaiser, and Polosukhin}]{vaswani2017attention}
Ashish Vaswani, Noam Shazeer, Niki Parmar, Jakob Uszkoreit, Llion Jones,
  Aidan~N Gomez, {\L}ukasz Kaiser, and Illia Polosukhin. 2017.
\newblock Attention is all you need.
\newblock In \emph{Advances in neural information processing systems}, pages
  5998--6008.

\bibitem[{Xie et~al.(2018)Xie, Sun, Huang, Tu, and Murphy}]{xie2018rethinking}
Saining Xie, Chen Sun, Jonathan Huang, Zhuowen Tu, and Kevin Murphy. 2018.
\newblock Rethinking spatiotemporal feature learning: Speed-accuracy trade-offs
  in video classification.
\newblock In \emph{Proceedings of the European Conference on Computer Vision
  (ECCV)}, pages 305--321.

\bibitem[{Xu et~al.(2016)Xu, Mei, Yao, and Rui}]{xu2016msr}
Jun Xu, Tao Mei, Ting Yao, and Yong Rui. 2016.
\newblock Msr-vtt: A large video description dataset for bridging video and
  language.
\newblock In \emph{Proceedings of the IEEE conference on computer vision and
  pattern recognition}, pages 5288--5296.

\bibitem[{Yu et~al.(2018)Yu, Kim, and Kim}]{yu2018joint}
Youngjae Yu, Jongseok Kim, and Gunhee Kim. 2018.
\newblock A joint sequence fusion model for video question answering and
  retrieval.
\newblock In \emph{Proceedings of the European Conference on Computer Vision
  (ECCV)}, pages 471--487.

\bibitem[{Zhou et~al.(2020)Zhou, Palangi, Zhang, Hu, Corso, and
  Gao}]{zhou2020unified}
Luowei Zhou, Hamid Palangi, Lei Zhang, Houdong Hu, Jason~J Corso, and Jianfeng
  Gao. 2020.
\newblock Unified vision-language pre-training for image captioning and vqa.
\newblock In \emph{AAAI}, pages 13041--13049.

\bibitem[{Zhou et~al.(2017)Zhou, Xu, and Corso}]{zhou2017towards}
Luowei Zhou, Chenliang Xu, and Jason~J Corso. 2017.
\newblock Towards automatic learning of procedures from web instructional
  videos.
\newblock \emph{arXiv preprint arXiv:1703.09788}.

\bibitem[{Zhou et~al.(2018)Zhou, Zhou, Corso, Socher, and Xiong}]{zhou2018end}
Luowei Zhou, Yingbo Zhou, Jason~J Corso, Richard Socher, and Caiming Xiong.
  2018.
\newblock End-to-end dense video captioning with masked transformer.
\newblock In \emph{Proceedings of the IEEE Conference on Computer Vision and
  Pattern Recognition}, pages 8739--8748.

\bibitem[{Zhu and Yang(2020)}]{zhu2020actbert}
Linchao Zhu and Yi~Yang. 2020.
\newblock Actbert: Learning global-local video-text representations.
\newblock In \emph{Proceedings of the IEEE/CVF Conference on Computer Vision
  and Pattern Recognition}, pages 8746--8755.

\bibitem[{Zhukov et~al.(2019)Zhukov, Alayrac, Cinbis, Fouhey, Laptev, and
  Sivic}]{zhukov2019cross}
Dimitri Zhukov, Jean-Baptiste Alayrac, Ramazan~Gokberk Cinbis, David Fouhey,
  Ivan Laptev, and Josef Sivic. 2019.
\newblock Cross-task weakly supervised learning from instructional videos.
\newblock In \emph{Proceedings of the IEEE/CVF Conference on Computer Vision
  and Pattern Recognition}, pages 3537--3545.

\end{thebibliography}

\clearpage
\appendix

\begin{table*}[!t]
\centering
\scalebox{0.57}{
    \begin{tabular}{l|l}
    \hline
Query & Text of Top-1 video \\
\hline
\multicolumn{2}{l}{Objects (26\%)}\\
\hline
\textbf{cartoon} show for kids & \textbf{pokemon video game} play\\
little pet shop \textbf{cat} getting a bath and washed with little brush & several \textbf{dogs} playing dead\\
\hline
\multicolumn{2}{l}{Attributes of Objects (6\%)}\\
\hline
a little \textbf{boy} singing in front of judges and crowd & a \textbf{woman} singing on the voice\\
a woman is mixing \textbf{food} in a mixing bowl & a man is stirring \textbf{something} in a pot\\
\hline
\multicolumn{2}{l}{Action (6\%)}\\
\hline
a person is \textbf{connecting} \textbf{something} to system & a man \textbf{looks at} the \textbf{battery} of a computer\\
a boy \textbf{plays} grand theft auto 5 & a narrator \textbf{explains where to find} a rare vehicle in grand theft auto\\
a man is \textbf{giving a review} on a vehicle & a person is \textbf{discussing} a car\\
% a man speaks to children in a classroom & a man speaks to children in a classroom\\
% one micky mouse is talking to other & one micky mouse is talking to other\\
a \textbf{naked child} runs through a field & the \textbf{girl} shows the \textbf{boys} her medal in this cartoon\\
a man is singing and standing in the road & a man in sunglasses and a blue shirt beat boxes\\
% fireworks are being lit and exploding in a night sky & fireworks are being lit and exploding in a night sky\\
\hline
\multicolumn{2}{l}{Specific vs General (62\%)}\\
\hline
some cartoon characters are moving around an area & a cartoon girl and animal jumping on body of male guy girl image still shown displaying on screen\\
\textbf{baseball player} hits ball & \textbf{people} are playing baseball\\
% a rocket is lauching into a blue sky smoke is emerging from the base of the rocket & a rocket is lauching into a blue sky smoke is emerging from the base of the rocket\\
the man in the video is showing a brief viewing of how the movie is starting & scrolling the the menu of movieclips with different movie trailers\\
a \textbf{student} explains to his \textbf{teacher} about the sheep of another student & there is a \textbf{guy} talking to his \textbf{father}\\
a video about different sports & a woman talks about horse racing\\
\hline
    \end{tabular}
}
\caption{Error analysis for text-video retrieval of MSR-VTT on 100 errors: we group errors in four (4) categories: objects, attributes of objects, actions, and specific vs general. Specific videos for general queries (or vice versa) sometimes may not be errors but hard to evaluate. 
}
\label{tbl:err_vtt}
\end{table*}

\begin{table*}
\centering
\scalebox{0.85}{
    \begin{tabular}{l|l}
    \hline
Hypothesis & Reference \\
\hline
add the lamb to the \textcolor{red}{\textbf{pan}} & add the lamb to the \textbf{pot} \\
add the cilantro \st{cilantro} and lime juice to \underline{the pot} & cut the cilantro and lime\\
add the \textcolor{red}{\textbf{onions}} to a pot of water & add \textbf{flour} to the pot and stir\\
dip the \textcolor{red}{\textbf{onion rings}} into the batter & dip the \textbf{shrimp} in the batter\\
add water to the bowl and mix & pour water into the \textbf{flour mixture} and mix\\
% add the tofu and water to the pan & add the cut tofu in and add some water and stir\\
remove the \textcolor{red}{\textbf{mussels}} from the pot & once the \textbf{shrimps} are defrosted drain the water\\
% add the vegetables to the pan and stir & add shredded cabbage julienned carrots salt and pepper\\
% fold the wrapper over & roll up the spring roll\\
% fry the chicken in the oil & place the chicken into a pot of hot oil\\
add the sauce to the \textcolor{red}{\textbf{pan}} and stir & add the sauce to the \textbf{wok} and stir\\
add lemon juice to the pan and stir & add \textbf{rice vinegar} and lemon juice to the pan and stir\\
add the beef to the pan and stir & add the diced beef meat to it and roast it\\
\hline
    \end{tabular}
}
\caption{Error analysis for video captioning on Youcook2: VLM tends to make mistakes in recognizing objects of similar shapes and colors to generate the wrong text.}
\label{tbl:error_youcookcap}
\end{table*}

\begin{figure*}
\begin{subfigure}[t]{3.1in}
    \centering
    \includegraphics[width=2.1in]{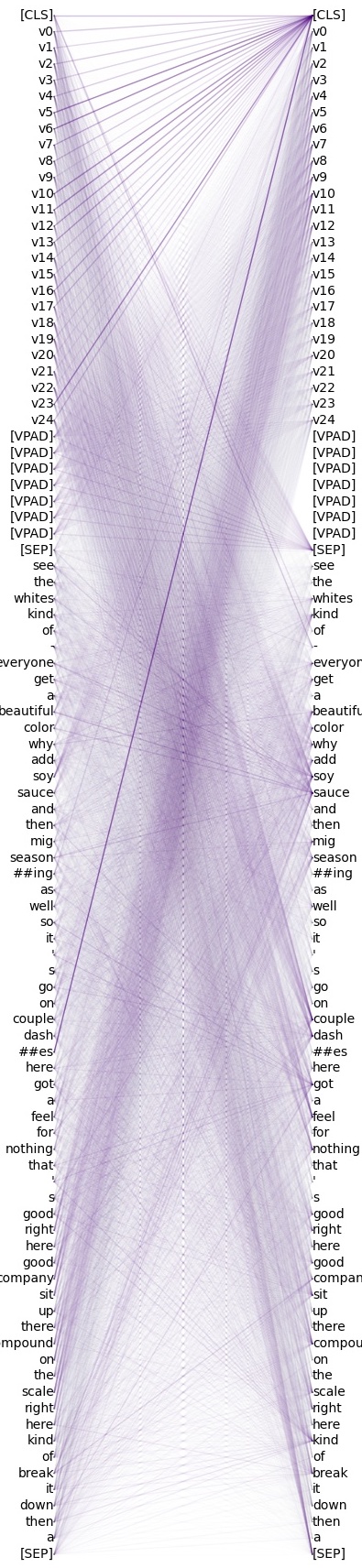}
    \caption{Layer 1, Head 1}
    \vspace{-3mm}
    \end{subfigure}
    \hfill
    \begin{subfigure}[b]{3.1in}
    \centering
    \includegraphics[width=2.1in]{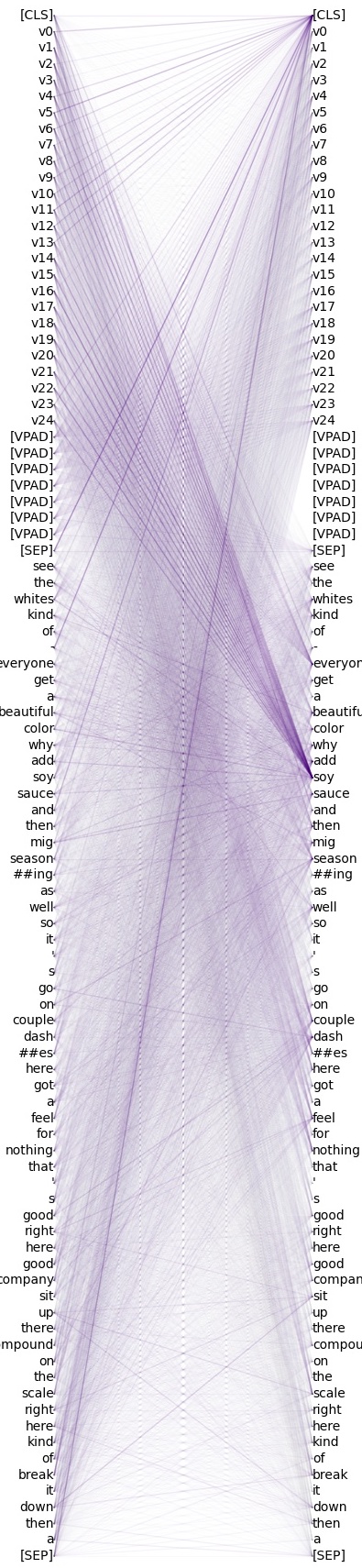}
    \caption{Layer 1, Head 5}
    \vspace{-3mm}
\end{subfigure}
\caption{Self-attention for video HfIeQ9pzL5U from 4:03 to 4:28: darker color indicates higher weights; v0-v24 are video tokens of 25 seconds.}
\label{fig:attn0_1}
\end{figure*}

\end{document}